\documentclass[conference]{IEEEtran}
\IEEEoverridecommandlockouts
\usepackage{amsmath,amssymb,amsfonts}
\usepackage{algorithmic}
\usepackage{hyperref}
\usepackage{graphicx}
\usepackage{textcomp}
\usepackage{xcolor}
\usepackage{acronym}
\usepackage{float}
\usepackage{multirow}
\usepackage{makecell}

\usepackage[backend=biber, sorting=none, style=numeric-comp, maxnames=5]{biblatex}

\acrodef{AC}[AC]{Coefficient of Variation}
\acrodef{ADC}[ADC]{Analog-to-Digital Converter}
\acrodef{ADEXP}[AdExp-I\&F]{Adaptive-Exponential Integrate and Fire}
\acrodef{ADM}[ADM]{Asynchronous Delta Modulator}
\acrodef{AER}[AER]{Address-Event Representation}
\acrodef{AEX}[AEX]{AER EXtension board}
\acrodef{AE}[AE]{Address-Event}
\acrodef{AFM}[AFM]{Atomic Force Microscope}
\acrodef{AGC}[AGC]{Automatic Gain Control}
\acrodef{AI}[AI]{Artificial Intelligence}
\acrodef{AMDA}[AMDA]{AER Motherboard with D/A converters}
\acrodef{AMPA}[AMPA]{$\alpha$-amino-3-hydroxy-5-methyl-4-isoxazolepropionic acid}
\acrodef{ANN}[ANN]{Artificial Neural Network}
\acrodef{API}[API]{Application Programming Interface}
\acrodef{APMOM}[APMOM]{Alternate Polarity Metal On Metal}
\acrodef{ARM}[ARM]{Advanced RISC Machine}
\acrodef{ASIC}[ASIC]{Application Specific Integrated Circuit}
\acrodef{AdExp}[AdExp-IF]{Adaptive Exponential Integrate-and-Fire}
\acrodef{BCM}[BMC]{Bienenstock-Cooper-Munro}
\acrodef{BD}[BD]{Bundled Data}
\acrodef{BEOL}[BEOL]{Back-end of Line}
\acrodef{BG}[BG]{Bias Generator}
\acrodef{BMI}[BMI]{Brain-Machince Interface}
\acrodef{BPTT}[BPTT]{Backpropagation through time}
\acrodef{BTB}[BTB]{band-to-band tunnelling}
\acrodef{CAD}[CAD]{Computer Aided Design}
\acrodef{CAM}[CAM]{Content Addressable Memory}
\acrodef{CAVIAR}[CAVIAR]{Convolution AER Vision Architecture for Real-Time}
\acrodef{CA}[CA]{Cortical Automaton}
\acrodef{CCN}[CCN]{Cooperative and Competitive Network}
\acrodef{CDR}[CDR]{Clock-Data Recovery}
\acrodef{CFC}[CFC]{Current to Frequency Converter}
\acrodef{CHP}[CHP]{Communicating Hardware Processes}
\acrodef{CMIM}[CMIM]{Metal-insulator-metal Capacitor}
\acrodef{CML}[CML]{Current Mode Logic}
\acrodef{CMOL}[CMOL]{Hybrid CMOS nanoelectronic circuits}
\acrodef{CMOS}[CMOS]{complementary metal–oxide–semiconductor}
\acrodef{CNN}[CNN]{Convolutional Neural Network}
\acrodef{CNS}[CNS]{central nervous system}
\acrodef{COTS}[COTS]{Commercial Off-The-Shelf}
\acrodef{CPG}[CPG]{Central Pattern Generator}
\acrodef{CPLD}[CPLD]{Complex Programmable Logic Device}
\acrodef{CPU}[CPU]{Central Processing Unit}
\acrodef{CSM}[CSM]{Cortical State Machine}
\acrodef{CSP}[CSP]{Constraint Satisfaction Problem}
\acrodef{CTXCTL}[CTXCTL]{CortexControl}
\acrodef{CV}[CV]{Coefficient of Variation}
\acrodef{DAC}[DAC]{Digital to Analog Converter}
\acrodef{DAS}[DAS]{Dynamic Auditory Sensor}
\acrodef{DAVIS}[DAVIS]{Dynamic and Active Pixel Vision Sensor}
\acrodef{DBN}[DBN]{Deep Belief Network}
\acrodef{DBS}[DBS]{Deep Brain Stimulation}
\acrodef{DFA}[DFA]{Deterministic Finite Automaton}
\acrodef{DIBL}[DIBL]{drain-induced-barrier-lowering}
\acrodef{DI}[DI]{delay insensitive}
\acrodef{DMA}[DMA]{Direct Memory Access}
\acrodef{DNF}[DNF]{Dynamic Neural Field}
\acrodef{DNN}[DNN]{Deep Neural Network}
\acrodef{DOF}[DOF]{Degrees of Freedom}
\acrodef{DPE}[DPE]{Dynamic Parameter Estimation}
\acrodef{DPI}[DPI]{Differential Pair Integrator}
\acrodef{DRAM}[DRAM]{Dynamic Random Access Memory}
\acrodef{DRRZ}[DR-RZ]{Dual-Rail Return-to-Zero}
\acrodef{DR}[DR]{Dual Rail}
\acrodef{DSP}[DSP]{Digital Signal Processor}
\acrodef{DVS}[DVS]{Dynamic Vision Sensor}
\acrodef{DYNAP}[DYNAP]{Dynamic Neuromorphic Asynchronous Processor}
\acrodef{EBL}[EBL]{Electron Beam Lithography}
\acrodef{ECG}[ECG]{Electrocardiography}
\acrodef{ECoG}[ECoG]{Electrocorticography}
\acrodef{EDVAC}[EDVAC]{Electronic Discrete Variable Automatic Computer}
\acrodef{EEG}[EEG]{Electroencephalography}
\acrodef{EIN}[EIN]{Excitatory-Inhibitory Network}
\acrodef{EMG}[EMG]{Electromyography}
\acrodef{EM}[EM]{Expectation Maximization}
\acrodef{EOG}[EOG]{Electrooculogram}
\acrodef{EPSC}[EPSC]{Excitatory Post-Synaptic Current}
\acrodef{EPSP}[EPSP]{Excitatory Post-Synaptic Potential}
\acrodef{EZ}[EZ]{Epileptogenic Zone}
\acrodef{FDSOI}[FDSOI]{Fully-Depleted Silicon on Insulator}
\acrodef{FET}[FET]{Field-Effect Transistor}
\acrodef{FFT}[FFT]{Fast Fourier Transform}
\acrodef{FI}[F-I]{Frequency-Current}
\acrodef{FMA}[FMA]{Floating microelectrode array} 
\acrodef{FNN}[FNN]{Feed-forward Neural Network}
\acrodef{FPGA}[FPGA]{Field Programmable Gate Array}
\acrodef{FR}[FR]{Fast Ripple}
\acrodef{FSA}[FSA]{Finite State Automaton}
\acrodef{FSM}[FSM]{Finite State Machine}
\acrodef{GABA}[GABA]{$\gamma$-aminobutanoic acid}
\acrodef{GIDL}[GIDL]{gate-induced-drain-leakage}
\acrodef{GOPS}[GOPS]{Giga-Operations per Second}
\acrodef{GPU}[GPU]{Graphical Processing Unit}
\acrodef{GT}[GT]{Ground Truth}
\acrodef{GUI}[GUI]{Graphical User Interface}
\acrodef{HAL}[HAL]{Hardware Abstraction Layer}
\acrodef{HFO}[HFO]{High Frequency Oscillation}
\acrodef{HH}[H\&H]{Hodgkin \& Huxley}
\acrodef{HMM}[HMM]{Hidden Markov Model}
\acrodef{HRS}[HRS]{High-Resistive State}
\acrodef{HR}[HR]{Human Readable}
\acrodef{HSE}[HSE]{Handshaking Expansion}
\acrodef{HW}[HW]{Hardware}
\acrodef{ICT}[ICT]{Information and Communication Technology}
\acrodef{IC}[IC]{Integrated Circuit}
\acrodef{IF2DWTA}[IF2DWTA]{Integrate \& Fire 2--Dimensional WTA}
\acrodef{IFSLWTA}[IFSLWTA]{Integrate \& Fire Stop Learning WTA}
\acrodef{IF}[I\&F]{Integrate-and-Fire}
\acrodef{IMU}[IMU]{Inertial Measurement Unit}
\acrodef{INCF}[INCF]{International Neuroinformatics Coordinating Facility}
\acrodef{INI}[INI]{Institute of Neuroinformatics}
\acrodef{IO}[I/O]{Input/Output}
\acrodef{IPSC}[IPSC]{Inhibitory Post-Synaptic Current}
\acrodef{IPSP}[IPSP]{Inhibitory Post-Synaptic Potential}
\acrodef{IP}[IP]{Intellectual Property}
\acrodef{ISI}[ISI]{Inter-Spike Interval}
\acrodef{IoT}[IoT]{Internet of Things}
\acrodef{ITL}[ITL]{In-The-Loop}
\acrodef{JFLAP}[JFLAP]{Java - Formal Languages and Automata Package}
\acrodef{LEDR}[LEDR]{Level-Encoded Dual-Rail}
\acrodef{LFP}[LFP]{Local Field Potential}
\acrodef{LIFE}[LIFE]{Longitudinal Intrafascicular Electrodes}
\acrodef{LIF}[LI\&F]{Leak Integrate-and-Fire}
\acrodef{LLC}[LLC]{Low Leakage Cell}
\acrodef{LNA}[LNA]{Low-Noise Amplifier}
\acrodef{LPF}[LPF]{Low Pass Filter}
\acrodef{LRS}[LRS]{Low-Resistive State}
\acrodef{LSM}[LSM]{Liquid State Machine}
\acrodef{LTD}[LTD]{Long Term Depression}
\acrodef{LTI}[LTI]{Linear Time-Invariant}
\acrodef{LTP}[LTP]{Long Term Potentiation}
\acrodef{LTU}[LTU]{Linear Threshold Unit}
\acrodef{LUT}[LUT]{Look-Up Table}
\acrodef{LVDS}[LVDS]{Low Voltage Differential Signaling}
\acrodef{MCMC}[MCMC]{Markov-Chain Monte Carlo}
\acrodef{MEA}[MEA]{Multielectrode Arrays}
\acrodef{MEMS}[MEMS]{Micro Electro Mechanical System}
\acrodef{MFR}[MFR]{Mean Firing Rate}
\acrodef{MIM}[MIM]{Metal Insulator Metal}
\acrodef{MLP}[MLP]{Multilayer Perceptron}
\acrodef{ML}[ML]{Machine Learning}
\acrodef{MOSCAP}[MOSCAP]{Metal Oxide Semiconductor Capacitor}
\acrodef{MOSFET}[MOSFET]{Metal Oxide Semiconductor Field-Effect Transistor}
\acrodef{MOS}[MOS]{Metal Oxide Semiconductor}
\acrodef{MRI}[MRI]{Magnetic Resonance Imaging}
\acrodef{NCS}[NCS]{Neuromorphic Cognitive Systems}
\acrodef{NDFSM}[NDFSM]{Non-deterministic Finite State Machine} 
\acrodef{ND}[ND]{Noise-Driven}
\acrodef{NEF}[NEF]{Neural Engineering Framework}
\acrodef{NHML}[NHML]{Neuromorphic Hardware Mark-up Language}
\acrodef{NIL}[NIL]{Nano-Imprint Lithography}
\acrodef{NI}[NI]{Neural Interface}
\acrodef{NMDA}[NMDA]{N-Methyl-D-Aspartate}
\acrodef{NME}[NE]{Neuromorphic Engineering}
\acrodef{NN}[NN]{Neural Network}
\acrodef{NOC}[NoC]{Network-on-Chip}
\acrodef{NRZ}[NRZ]{Non-Return-to-Zero}
\acrodef{NSM}[NSM]{Neural State Machine}
\acrodef{OR}[OR]{Operating Room}
\acrodef{OTA}[OTA]{Operational Transconductance Amplifier}
\acrodef{PCB}[PCB]{Printed Circuit Board}
\acrodef{PCHB}[PCHB]{Pre-Charge Half-Buffer}
\acrodef{PCM}[PCM]{Phase Change Memory}
\acrodef{PCA}[PCA]{Personal Component Analysis}

\acrodef{PC}[PC]{Personal Computer}
\acrodef{PE}[PE]{Phase Encoding}
\acrodef{PFA}[PFA]{Probabilistic Finite Automaton}
\acrodef{PFC}[PFC]{prefrontal cortex}
\acrodef{PFM}[PFM]{Pulse Frequency Modulation}
\acrodef{PNI}[PNI]{peripheral nerve interface}
\acrodef{PNS}[PNS]{peripheral nervous system}
\acrodef{PPG}[PPG]{Photoplethysmography}
\acrodef{PR}[PR]{Production Rule}
\acrodef{PSC}[PSC]{Post-Synaptic Current}
\acrodef{PSP}[PSP]{Post-Synaptic Potential}
\acrodef{PSTH}[PSTH]{Peri-Stimulus Time Histogram}
\acrodef{PV}[PV]{Parvalbumin}
\acrodef{QDI}[QDI]{Quasi Delay Insensitive}
\acrodef{RAM}[RAM]{Random Access Memory}
\acrodef{RA}[RA]{Resected Area}
\acrodef{RDF}[RDF]{random dopant fluctuation}
\acrodef{RELU}[ReLu]{Rectified Linear Unit}
\acrodef{RLS}[RLS]{Recursive Least-Squares}
\acrodef{RMSE}[RMSE]{Root Mean Squared-Error}
\acrodef{RMS}[RMS]{Root Mean Squared}
\acrodef{RNN}[RNN]{Recurrent Neural Networks}
\acrodef{RNN}[RNN]{Recurrent Neural Network}
\acrodef{ROLLS}[ROLLS]{Reconfigurable On-Line Learning Spiking}
\acrodef{RRAM}[R-RAM]{Resistive Random Access Memory}
\acrodef{RSA}[RSA]{Respiratory Sinus Arrhythmia}
\acrodef{R}[R]{Ripples}
\acrodef{SAC}[SAC]{Selective Attention Chip}
\acrodef{SAT}[SAT]{Boolean Satisfiability Problem}
\acrodef{SCI}[SCI]{Spinal Cord Injury}
\acrodef{SCX}[SCX]{Silicon CorteX}
\acrodef{SD}[SD]{Signal-Driven}
\acrodef{SEM}[SEM]{Spike-based Expectation Maximization}
\acrodef{SLAM}[SLAM]{Simultaneous Localization and Mapping}
\acrodef{SNN}[SNN]{Spiking Neural Network}
\acrodef{SNR}[SNR]{Signal to Noise Ratio}
\acrodef{SOC}[SOC]{System-On-Chip}
\acrodef{SOI}[SOI]{Silicon on Insulator}
\acrodef{SOZ}[SOZ]{Seizure Onset Zone}
\acrodef{SP}[SP]{Separation Property}
\acrodef{SRAM}[SRAM]{Static Random Access Memory}
\acrodef{PYR}[PYR]{Pyramidal}
\acrodef{SST}[SST]{Somatostatin}

\acrodef{STDP}[STDP]{Spike-Timing Dependent Plasticity}
\acrodef{STD}[STD]{Short-Term Depression}
\acrodef{STP}[STP]{Short-Term Plasticity}
\acrodef{STT-MRAM}[STT-MRAM]{Spin-Transfer Torque Magnetic Random Access Memory}
\acrodef{STT}[STT]{Spin-Transfer Torque}
\acrodef{SVM}[SVM]{Support Vector Machine}
\acrodef{SW}[SW]{Software}
\acrodef{TCAM}[TCAM]{Ternary Content-Addressable Memory}
\acrodef{TFT}[TFT]{Thin Film Transistor}
\acrodef{TIME}[TIME]{Transverse Intrafascicular Multichannel Electrode}
\acrodef{TLE}[TLE]{Temporal Lobe Epilepsy}
\acrodef{UEA}[UEA]{Utah electrode array}
\acrodef{USB}[USB]{Universal Serial Bus}
\acrodef{USEA}[USEA]{Utah Slanted Electrode Array}
\acrodef{VHDL}[VHDL]{VHSIC Hardware Description Language}
\acrodef{VIP}[VIP]{Vasoactive Intestinal Peptide}
\acrodef{VLSI}[VLSI]{Very Large Scale Integration}
\acrodef{VNS}[VNS]{Vagal Nerve Stimulation}
\acrodef{VOR}[VOR]{Vestibulo-Ocular Reflex}
\acrodef{WCST}[WCST]{Wisconsin Card Sorting Test}
\acrodef{WTA}[WTA]{Winner-Take-All}
\acrodef{XML}[XML]{eXtensible Mark-up Language}
\acrodef{divmod3}[DIVMOD3]{divisibility of a number by three}
\acrodef{hWTA}[hWTA]{hard Winner-Take-All}
\acrodef{iEEG}[iEEG]{intracranial electroencephalography}
\acrodef{sWTA}[sWTA]{soft Winner-Take-All}

\DeclareSourcemap{
  \maps[datatype=bibtex]{
    \map[overwrite=true]{
      \step[fieldset=url, null]
      \step[fieldset=eprint, null]
    }
  }
}

\newcommand{\appropto}{\mathrel{\vcenter{
  \offinterlineskip\halign{\hfil$##$\cr
    \propto\cr\noalign{\kern2pt}\sim\cr\noalign{\kern-2pt}}}}}
    
\def\BibTeX{{\rm B\kern-.05em{\sc i\kern-.025em b}\kern-.08em
    T\kern-.1667em\lower.7ex\hbox{E}\kern-.125emX}}
\begin{document}

\title{Mixed-signal implementation of feedback-control optimizer for single-layer Spiking Neural Networks}

\author{\IEEEauthorblockN{Jonathan Haag\IEEEauthorrefmark{1}\IEEEauthorrefmark{2},
Christian Metzner\IEEEauthorrefmark{1}\IEEEauthorrefmark{2},
Dmitrii Zendrikov\IEEEauthorrefmark{2},
Giacomo Indiveri\IEEEauthorrefmark{2},~\IEEEmembership{Senior Member, IEEE},\\
Benjamin Grewe\IEEEauthorrefmark{2}\IEEEauthorrefmark{3},
Chiara De Luca\IEEEauthorrefmark{2}\IEEEauthorrefmark{4}\IEEEauthorrefmark{5}, and
Matteo Saponati\IEEEauthorrefmark{2}\IEEEauthorrefmark{5}}
\IEEEauthorblockA{\IEEEauthorrefmark{1} equal contribution, \IEEEauthorrefmark{5} equal contribution}
\IEEEauthorblockA{\IEEEauthorrefmark{2}Institute of Neuroinformatics, University of Zürich and ETH Zürich, Zürich, Switzerland}
\IEEEauthorblockA{\IEEEauthorrefmark{3}AI Center, ETH, Zürich, Switzerland}
\IEEEauthorblockA{\IEEEauthorrefmark{4}Digital Society Initiative, University of Zurich, Switzerland}
%
\thanks{}}

\maketitle

\begin{abstract}
On-chip learning is key to scalable and adaptive neuromorphic systems, yet existing training methods are either difficult to implement in hardware or overly restrictive.
However, recent studies show that feedback-control optimizers can enable expressive, on-chip training of neuromorphic devices.
In this work, we present a proof-of-concept implementation of such feedback-control optimizers on a mixed-signal neuromorphic processor. 
We assess the proposed approach in an In-The-Loop (ITL) training setup on both a binary classification task and the nonlinear Yin–Yang problem, demonstrating on-chip training that matches the performance of numerical simulations and gradient-based baselines. 
Our results highlight the feasibility of feedback-driven, online learning under realistic mixed-signal constraints, and represent a co-design approach toward embedding such rules directly in silicon for autonomous and adaptive neuromorphic computing.
\end{abstract}


\section{Introduction}
\label{sec:introduction}
\acp{SNN} are increasingly studied as a biologically inspired and hardware-efficient alternative to artificial neural networks (ANN). Their event-driven dynamics make them particularly well-suited for neuromorphic systems, which exploit mixed-signal circuits to emulate spiking neurons in real time~\cite{chiccaNeuromorphicElectronicCircuits2014, moradiScalableMulticoreArchitecture2018}. A major challenge, however, remains the development of scalable and hardware-compatible learning rules. Conventional approaches such as \ac{BPTT}~\cite{bohteErrorbackpropagationTemporallyEncoded2002} achieve high accuracy in software but are inherently non-local: they require information from the entire network (non-local in space) and must process sequences in batches while storing activity over time (non-local in time). These requirements make direct deployment on neuromorphic substrates highly impractical.  

In contrast, recent work has focused on local and online learning rules. Online learning refers to weight updates that are triggered continuously by the occurrence of individual spikes, without waiting for an entire batch or full sequence to be processed. This property not only reduces memory requirements but also allows learning to proceed in real time. The importance extends beyond algorithmic elegance: it is critical for enabling subject-specific adaptation, online fine-tuning, and task-dependent optimization in practical deployments. Saponati \emph{et al.}~\cite{saponati2025feedback} introduced the spike-based feedback control algorithm, which encodes error signals directly in spike trains and drives local synaptic plasticity. Such rules represent a promising route to neuromorphic learning, as they avoid dependence on global error information and enable continuous updates using signals naturally available on hardware.  
However, deploying such learning rules on neuromorphic substrates remains non-trivial. Mixed-signal devices exhibit mismatch, noise, and quantization constraints, while internal state variables are only partially observable, making neuromorphic hardware both a demanding and realistic testbed for the robustness of spike-based learning methods.  

In this work, we use the DYNAP-SE neuromorphic processor~\cite{moradiScalableMulticoreArchitecture2018} to provide a proof of concept of the spike-based feedback control algorithm. The DYNAP-SE processor is a mixed-signal platform in which analog circuits exploit device physics to emulate spiking neuron dynamics. The chip consists of multiple cores of \ac{ADEXP} neurons interconnected via asynchronous address-event routing, operating in real time and supporting plastic synaptic connections. To co-design an on-chip implementation of such an algorithm, we employ a computer-in-the-loop training procedure, in which a host computer computes weight updates and reconfigures synapses while the chip executes network dynamics. This approach allows us to assess the algorithm under realistic hardware constraints, while paving the way toward future neuromorphic systems that embed the learning rule directly in silicon.
\section{Methods}
\label{sec:methods}
\subsection{Spiking feedback-control optimizer}
\begin{figure}
    \centering
    \includegraphics[width=0.7\columnwidth]{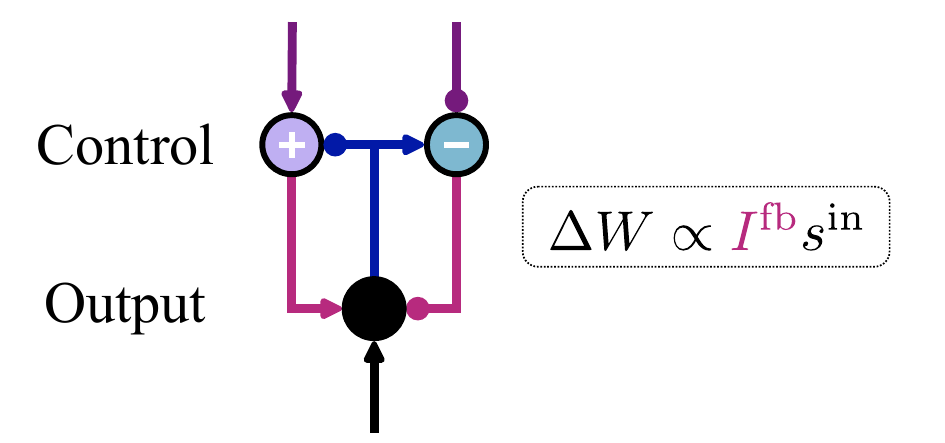}
    \vspace*{-4mm}
    \caption{Schematic of the  computational primitive. One output neuron (black) receives inputs (black arrow) and sends output (blue arrows) to the two corresponding positive (purple) and negative (cyan) control neurons. Arrows and circles indicate excitation and inhibition, respectively.
    }\label{fig:control-sketch}
\end{figure}

We adopt the feedback-control optimizer introduced in~\cite{saponati2025feedback} to train single-layer \acp{SNN} on Neuromorphic devices. 
The architecture is composed of a Leaky Integrate-and-Fire (LIF) output neuron per class, with two additional controller neurons per output (positive and a negative).
These neurons receive target and output spike trains and recurrently project back to the output neuron (Fig.~\ref{fig:control-sketch}). 
By construction, the positive control neuron increases the activity of the output neurons when the target rate exceeds the actual firing rate, while the negative control neuron decreases it when the opposite holds. 
Together, they implement a proportional-integral spiking controller that encodes the error between the target and the output activity directly into spike-based feedback currents. 
The recurrent feedback currents provide a local learning signal to update synaptic weights online according to,%
\begin{equation}
    w_t = w_{t-1} + \eta I^{\mathrm{fb}}_{t} s^{\mathrm{in}}_{t},
    \label{eq:sfc_update}
\end{equation}
where $s^{\mathrm{in}}_t$ denotes the presynaptic input spikes, $I^{\mathrm{fb}}_{t}$ the feedback current generated by the controller and received by the neuron at timestep $t$, and $\eta$ the learning rate. 
This update rule is local, online, and hardware compatible, making it well-suited for implementation on mixed-signal neuromorphic chips.

\subsection{The DYNAP-SE chip}\label{sec:method_dynapse}
\begin{figure*}
    \centering
    \includegraphics[width=1\textwidth]{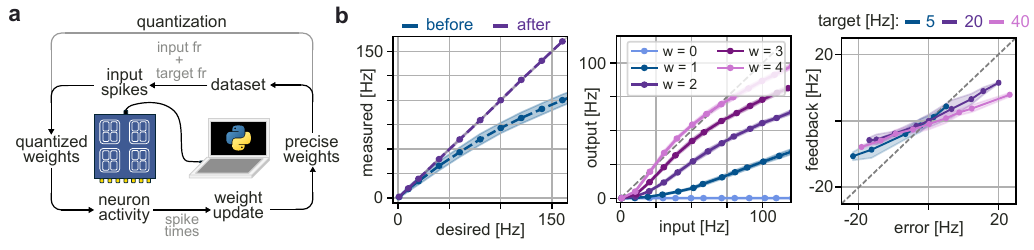}
    \vspace*{-9mm}
    \caption{\textbf{a}) In-the-loop (ITL) training setup.
    \textbf{b}
    Left: Calibration of input and target spike generation with recorder neurons, before and after calibration.
    Middle: Frequency-frequency curves of calibrated DYNAP-SE neurons for different synaptic weights (see legend).
    Right: relation between error (output activity minus target activity) and the total feedback (activity of the control neurons) in Hz.
    }
    \label{fig:diagram-calibration}
\end{figure*}

We test the feedback-control optimizer with the DYNAP-SE neuromorphic processor, a multi-core asynchronous mixed-signal system optimized for real-time emulation of spiking neural dynamics~\cite{Moradi_etal18}. 
The device comprises four processing cores, each embedding 256 current-mode silicon neurons~\cite{Livi_Indiveri09,Chicca_etal14} that implement adaptive exponential integrate-and-fire (\ac{ADEXP}) models~\cite{Brette_Gerstner05}. 

Every neuron supports up to 64 programmable synaptic inputs, configured through \ac{CAM} blocks that store the addresses of presynaptic sources. 
Synapses can be set as excitatory (positive weight) or inhibitory (negative weight), and operate with fast or slow dynamics.
Since the analog circuits implementing the synapses are inherently noisy and nonlinear, we instead encode synaptic strength by varying the number of parallel synaptic connections between neural units.
This effectively discretizes synaptic weights to a maximum resolution of 6 bits.

Neurons within a core share global parameters, including leak constants, time constants, and refractory periods, and each neuron provides up to 1024 fan-out connections. 
Inter-core and intra-core communication is realized through an \ac{AER} protocol~\cite{Deiss_etal98}, which guarantees microsecond-level timing precision under event-driven, high-load conditions. 
In the high-input regime and with adaptation disabled, the underlying circuit dynamics of an \ac{ADEXP} neuron reduce to
\begin{equation*}
\tau \frac{d}{dt} I_{\text{mem}} + I_{\text{mem}} \approx \frac{I_{\text{in}} I_{\text{gain}}}{I_{\tau}} + \frac{I_a I_{\text{mem}}}{I_{\tau}},
\end{equation*}
where $I_{\text{mem}}$ denotes the current representing the membrane potential, $\tau$ is the effective membrane time constant, and the second term on the right-hand side corresponds to the positive feedback loop characteristic of the \ac{ADEXP} model.
Due to device mismatch, \ac{ADEXP} neurons can exhibit variability within and across cores as well as being sensitive to signal noise. 
To address this, we use populations of 10 \ac{ADEXP} neurons to represent each neural unit.
Each unit's activity is the average firing rate across the entire population.
This population-based representation reduces sensitivity to neuron-to-neuron variability and noise fluctuations.

\subsection{In-the-loop training} 
\label{sec:itl_training}
We evaluate the feedback-control algorithm by training a DYNAP-SE processor on-chip with an \ac{ITL} training setup. 
In this configuration, the chip is interfaced with a host computer via USB and controlled through the \texttt{Samna} Python library~\cite{samna2025}. 
The host is responsible for the overall experiment orchestration (sending input spikes from the dataset, sending target spikes for classification, calculating the weight updates, and so on), while the DYNAP-SE performs the forward pass in real time.
Each training iteration proceeds as follows and as depicted in Figure~\ref{fig:diagram-calibration}a. First, inputs and target firing rates are converted to spike trains and sent to the chip. 
The network dynamics are implemented on-chip, and the output spikes are collected asynchronously through the \ac{AER} interface. 
After a fixed time window of duration $T$, the recorded spike times are transferred to the host. 
The host estimates the feedback current from the activity of the control neurons and calculates floating-point weight updates according to the learning rule in~\eqref{eq:sfc_update}. 
The updated weights are then quantized to the resolution supported by the hardware synapses and mapped back to the connection matrix of the DYNAP-SE. 
This procedure requires only adding or removing discrete synaptic connections; however, if a weight changes sign, the synapse type must also be switched from excitatory to inhibitory or vice versa, requiring careful calibration.
Because the DYNAP-SE runs continuously in real time, the duration of each training step directly corresponds to the wall-clock time (i.e., recording activities over $T$ milliseconds requires $T$ milliseconds of measurement). 
While this limits throughput, it ensures that the learning rule is tested under realistic hardware constraints. 
This \ac{ITL} approach thus provides a practical framework to assess the functionality and robustness of spike-based learning rules on mixed-signal neuromorphic substrates, before moving towards fully embedded on-chip adaptation.
\subsection{Datasets}
\label{sec:datasets}
We test the optimizer with two benchmark datasets, both following the definitions in~\cite{saponati2025feedback}. 
The first is a binary classification task for single-layer SNNs. 
Each example consists of two input spike trains: in class~A, the first input neuron fires at a high rate while the second fires at a lower rate, while the opposite is true for  class~B inputs. 
%
%
Input examples are presented for fixed-duration windows, after which synaptic weights are updated.
The second dataset is the Yin–Yang benchmark~\cite{krienerYinYangDataset2022}, a non-linearly separable, three-class problem. 
Each sample belongs to the Yin, Yang, or Dot region of the symbol and is encoded by four input features $(x, y, 1-x, 1-y)$. 
These coordinates are mapped to firing rates and converted to Poisson spike trains, as described in~\cite{saponati2025feedback}. 
In both cases, the output layer includes one neuron per class (two neurons for binary classification, three neurons for Yin-Yang), whose firing rates are modulated by the control neurons to reach high target activity for their corresponding class inputs and low activity for all others.
As in the binary task, each training example is presented once for a fixed duration, and separate training, validation, and test sets are generated randomly to ensure robustness of the evaluation.%
\section{Results}
\label{sec:results}
\subsection{Network calibration on the DYNAP-SE}
Accurate evaluation of the feedback control algorithm requires calibration of neuron, synapse, and controller parameters on the DYNAP-SE. 
Due to analog variability and limited observability of the system, we calibrate the neuron activities through targeted experiments, summarized in Fig.~\ref{fig:diagram-calibration}b.

Since FPGA-based virtual neurons cannot be monitored directly, we first calibrated the input and target spike generation using recorder neurons.
In particular, we route the output of the virtual neurons through the recorders neurons present on-chip and correct the mapping to significantly reduce the mismatch between desired and measured activities (Fig.~\ref{fig:diagram-calibration}b, left).
Second, we measure the synaptic transfer characteristics by driving pairs of neurons with controlled presynaptic firing rates and varying synaptic weight. The resulting frequency–frequency curves (Fig.~\ref{fig:diagram-calibration}b, middle) confirm that larger weights reliably increase postsynaptic rates, while highlighting the practical upper bound imposed by the 64-input fan-in constraint. We use this information to choose the effective weight range used during training. 
Finally, we tune the controller biases to align the differential activity of positive and negative control neurons with the target error signal. 
After calibration, the feedback signal $r^{\mathrm{fb}} = r^{+} - r^{-}$ linearly tracks the difference between output and target activity, i.e. the error signal (Fig.~\ref{fig:diagram-calibration}b, right). 
This step is critical to ensure that the feedback current used in learning reliably encodes the error magnitude and sign, up to a scaling factor that can be compensated by the learning rate.

Overall, these calibration procedures establish stable operating conditions on the DYNAP-SE for robust ITL training..

\subsection{Binary classification dataset}

\begin{figure*}
    \centering
    \includegraphics[width=0.98\textwidth]{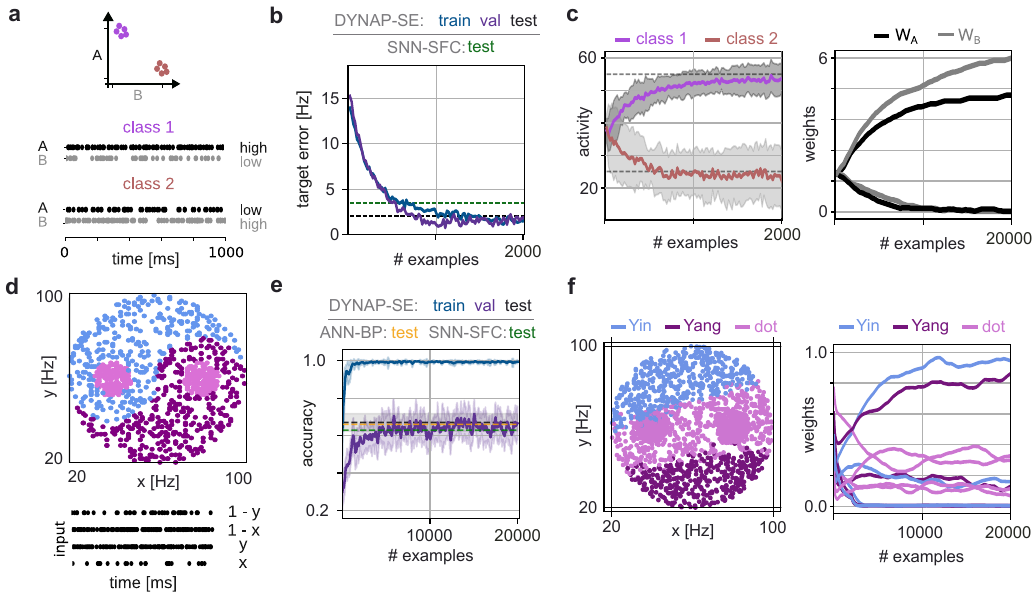}
    \vspace*{-5mm}
    \caption{\textbf{Successful ITL training with the DYNAP-SE.}
    \textbf{a}) Binary classification task. 
    \textbf{b}) Target error during ITL training and comparison with numerical simulations.
    SNN-SFC stands for Spiking Neural Network - Spiking Feedback Control.
    \textbf{c}) Left: Activity of the output neuron during ITL training when given inputs from class 1 and 2 (see legend).
    Right: Dynamics of the input weights.
    \textbf{d}) Poisson encoding of the Yin Yang dataset.
    \textbf{e}) Same as \textbf{b} for classification accuracy.
    ANN-BP stands for Artificial Neural Network - Backpropagation.
    \textbf{f}) Left: Network inference at test time.
    Right: Dynamics of the input weights.
    }
    \label{fig:results}
\end{figure*}

We first evaluate the feedback-control optimzier on the binary classification task (Fig.~\ref{fig:results}a). 
The absolute difference between the output and the target activity (i.e., the target error) decreases steadily during ITL training (Fig.~\ref{fig:results}b).
After training, the hardware implementation achieves \textbf{100\%} accuracy and a \textbf{2.1\,Hz} target error, closely matching corresponding numerical simulations (Table~\ref{tab:results}).
During the course of training, the network activity without the feedback drive gets closer to the target activities for both class 1 and 2 (Fig.~\ref{fig:results}c, left).
As the system learns to reach the targets without active feedback, the synaptic weights are adjusted incrementally, gradually diverging from their constant initialization (Fig.~\ref{fig:results}c, right).


\subsection{Non-linear Yin–Yang dataset}
Next, we evaluate the optimizer on the Yin–Yang benchmark, a three-class problem requiring non-linear decision boundaries (see Fig.~\ref{fig:results}d and Section~\ref{sec:datasets}). 
The network consists of four input neurons representing the dot coordinates $(x, y, 1-x, 1-y)$ and three output neurons. 
Input firing rates ranged from 20 to 100\,Hz, while target neurons were driven at 20\,Hz for the correct class and 2\,Hz otherwise. 

Figure~\ref{fig:results}e shows the training and validation accuracy during ITL training of the DYNAP-SE.
The hardware implementation achieves \textbf{67\%} accuracy at test time, closely matching the numerical simulations (\textbf{63\%}) and the ANN baseline (\textbf{66\%}) for a single linear layer, as reported in \cite{saponati2025feedback}. We estimated the power consumption during the inference phase on the DYNAP-SE chip~\cite{moradiScalableMulticoreArchitecture2018} for each experiment, resulting in an order of magnitude of $10-100 \mu W$.
Table~\ref{tab:results} summarizes these results.
Figure~\ref{fig:results}f shows test inference (without active feedback), revealing a clear separation of the three regions.
Furthermore, the synaptic weights of the different output neurons diverge from their random Gaussian initialization and stabilize after $\sim$10k training examples (Figure~\ref{fig:results}f, right).

Overall, the results on both classification tasks demonstrate successful training of single-layer SNNs directly on neuromorphic hardware using the spike-based learning rule of the feedback-control algorithm. 

\begin{table}[h!]
\centering
\caption{
Test-time accuracy for standard single-layer ANN: \textbf{100}\% on binary classification, \textbf{66}\% on Yin Yang \cite{saponati2025feedback}. Estimated power consumption on DYNAP-SE during inference.}
\label{tab:results}
\vspace*{-1.5mm}
\begin{tabular}{|l|l|c|c|c|}
\hline
\textbf{dataset} & \makecell{\textbf{test-time}\\
\textbf{metric}} & \makecell{\textbf{numerical}\\
\textbf{sim.}} & \textbf{DYNAP-SE}  & \makecell{\textbf{power}\\
\textbf{$[\mu W]$}}\\
\hline
\multirow{2}{*}{Binary} 
 & accuracy [\%] &  \textbf{100} & \textbf{100}  & \multirow{2}{*}{41}\\ 
 & target error\,[Hz]  & \textbf{3.9} & \textbf{2.1} &\\

\hline
\multirow{2}{*}{Yin–Yang}
 & accuracy [\%] &  \textbf{63} & \textbf{67}  & \multirow{2}{*}{189}\\ 
 & target error\,[Hz]  & \textbf{7.2} & \textbf{6.1}& \\
\hline
\end{tabular}
\end{table}
\section{Conclusions}
\label{sec:conclusions}
In this work, we demonstrated the implementation of a spike-based feedback control learning rule on the DYNAP-SE neuromorphic processor. 
Using an ITL setup, we were able to test the algorithm directly on mixed-signal hardware, accounting for device mismatch, noise, and limited synaptic resolution. 
The results on both binary and Yin–Yang classification tasks show that our approach achieves performance comparable to software simulations and gradient-based baselines, while operating fully in real time. 
The continuous, event-driven dynamics of the DYNAP-SE allowed weight updates to be applied rapidly and seamlessly during training, highlighting the suitability of the platform for exploring hardware–algorithm co-design.  
Because the learning rule is not yet implemented directly in silicon, a host computer was required to calculate weight updates and reconfigure synapses. 
Nonetheless, this computer-in-the-loop scheme enabled systematic evaluation under realistic hardware constraints and established a path toward embedded on-chip learning. 
Extending this framework to multi-layer networks requires rules for learning the feedback weights to hidden layers themselves, in addition to the rule introduced here for learning feedforward weights.
Backpropagation-free frameworks such as \cite{meulemansCreditAssignmentNeural2021, ororbiaBackpropagationFreeDeepLearning2023} provide useful starting points.
Furthermore, future work will include the design of custom neuromorphic processors embedding the feedback control rule natively in silicon. 
Such developments would eliminate the need for external supervision, paving the way for scalable and autonomous online learning in SNNs.

\section*{Acknowledgment}
M.S. and J.H. acknowledge the ETH Zürich Postdoctoral Fellowship (nr. 23-2 FEL-042).
C.D.L and J.H. were supported by the DSI at University of Zurich (grant no.G-95017-01-12). 
G.I. was supported by the HORIZON EUROPE EIC Pathfinder Grant ELEGANCE (Grant No. 101161114) and received funding from the Swiss State Secretariat for Education, Research and Innovation (SERI).
B.G. was supported by the Swiss National Science Foundation (CRSII5-173721, 315230 189251) and ETH project funding (24-2 ETH-032).

%

%


\vspace{\fill} 
\newpage
\printbibliography

\end{document}